\documentclass[11pt]{article}

\usepackage[final]{acl}

\usepackage{times}
\usepackage{latexsym}
\usepackage{booktabs}
\usepackage{adjustbox}
\usepackage{amsmath}
\usepackage{hyperref}

\usepackage[T1]{fontenc}

\usepackage[utf8]{inputenc}

\usepackage{microtype}

\usepackage{inconsolata}

\usepackage{graphicx}

\usepackage[table]{xcolor}
\usepackage{colortbl}

\definecolor{rankA}{HTML}{006837}
\definecolor{rankB}{HTML}{1A9850}
\definecolor{rankC}{HTML}{66BD63}
\definecolor{rankD}{HTML}{A6D96A}
\definecolor{rankE}{HTML}{D9EF8B}
\definecolor{rankF}{HTML}{FFFFBF}
\definecolor{rankG}{HTML}{FEE08B}
\definecolor{rankH}{HTML}{FDAE61}
\definecolor{rankI}{HTML}{F46D43}
\definecolor{rankJ}{HTML}{D73027}
\definecolor{rankK}{HTML}{A50026}

\title{Rethinking Pivot Programming Languages in Code Language Models}

\author{Andor Diera \\
  Ulm University\\
  Ulm, Germany\\
  \texttt{andor.diera@uni-ulm.de} \\\And
 Lukas Galke Poech \\
  University of
Southern Denmark\\
  Odense, Denmark\\
  \texttt{galke@imada.sdu.dk} \\\And
  Matthias Tichy \\
  Ulm University\\
  Ulm, Germany\\
  \texttt{matthias.tichy@uni-ulm.de} \\}

\usepackage{todonotes}
\begin{document}
\maketitle
\begin{abstract}

Multilingual code language models transfer skills across programming languages (PLs), but whether any PL occupies a privileged pivot position remains contested: geometric analyses point to C-family languages and Go, while behavioral evidence highlights Python. We revisit this question under controls for representational anisotropy and length variation across PLs, two confounds that compromise prior cosine-based analyses. Across three code models on multilingual competitive-programming data, we study three views of cross-PL organization: pairwise PL geometry, PL-English alignment, and pivoted retrieval through candidate PL representation spaces. The results are relation-dependent. Code-code geometry reveals structured language regions but no universal center; code-English alignment favors high-level scripting languages; and pivoted retrieval favors different intermediate spaces for code-to-code and English-to-code transfer. These findings suggest that Python’s special role is better understood as English-facing affinity than as universal geometric centrality\footnote{All code and scripts needed to reproduce our experiments can be found at \href{https://github.com/drndr/codelm_hub_lang}{github.com/drndr/codelm\_hub\_lang}}.

\end{abstract}

\section{Introduction}

Multilingual code language models (LMs) can translate between programming languages, generate code from natural language specifications, and transfer learned skills across languages~\cite{roziere2023code, hui2024qwen2, jiang2026large}. These behavioral capabilities are well established, yet the underlying internal mechanisms remain underexplored. What representational structure allows a single model to serve many languages, and does it privilege certain languages over others? For natural language, a growing body of interpretability work has converged on a surprising answer: multilingual LMs internally route through English-like representations regardless of input language, effectively using English as an implicit pivot~\cite{wendler2024llamas}. This semantic-hub structure has been shown to extend beyond natural language to other modalities including code, mathematics, and visual/audio inputs~\cite{wu2025semantic}.

\begin{figure}[t]
    \centering
    \includegraphics[width=\columnwidth]{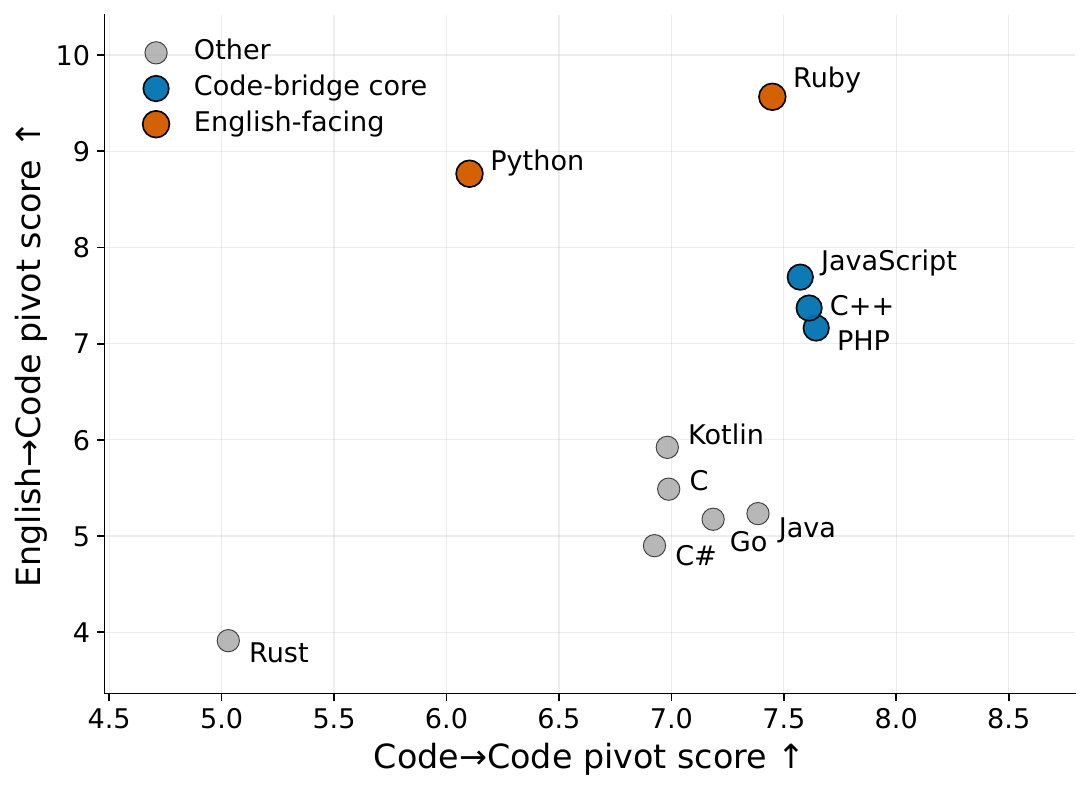}
    \caption{
    \textbf{Pivoted retrieval} split by transfer type. Each point is a candidate programming-language pivot, positioned by its mean within-task rank for code-to-code retrieval and English-to-code retrieval averaged across 3 code-LMs. For visualization, mean within-task pivot ranks are transformed into pivot scores, with higher values indicating better average rank.
    }
    \label{fig:functional-pivot-transfer-type}
\end{figure}

However, whether code LMs exhibit an analogous pivot toward a
particular \emph{programming language} (PL) remains contested.
Geometric analyses point in one direction:
\citet{kargaran2025programming} apply logit lens~\cite{belrose2023eliciting}
and MEXA cross-lingual alignment~\cite{kargaran2025mexa}, concluding
that English remains the implicit pivot in code LMs, with C-family
languages (C, C++, C\#, Java) most closely aligned among programming languages; 
\citet{yun2025beyond} additionally report Go as the geometric
centroid of code representation space. These cross-PL alignment
claims rest on representation-similarity measures vulnerable to two
confounds in transformer embedding spaces: representational
anisotropy~\cite{ethayarajh2019contextual, godey2024anisotropy,
DBLP:conf/esann/DieraGS25} and substantial character-length variation across
PLs, neither of which has been systematically controlled for.
Behavioral and training-dynamics evidence, by contrast, points in
another direction: \citet{moumoula2026programming} document Python
as a universal attractor in code generation, and
\citet{yang2026scaling} find that Python uniquely benefits other
languages when used as a pre-training auxiliary.

In this paper, we revisit the pivot-language question under explicit geometric controls. We analyze hidden-state representations from multilingual code models, comparing both code--code and code--English geometry. Rather than treating ``centrality'' as a single measure, we separate three questions: where languages sit in controlled centroid geometry, which languages align most closely with English descriptions, and which languages function best as intermediate spaces for downstream retrieval. To address anisotropy and length effects, we use mean-centered representations and length-matched code subsets. To assess whether candidate programming-language spaces can act as useful intermediates, we introduce a pivoted retrieval diagnostic. In this diagnostic, source and target representations are mapped into a candidate pivot space using Procrustes alignment, and the pivot is evaluated by same-problem retrieval ranking.

Our results show that the answer depends on the relation being bridged. Controlled code--code geometry reveals structured language regions, but does not identify a single universal PL center. Code--English geometry favors high-level scripting-like languages rather than C-family languages. Most importantly, pivoted retrieval splits by transfer type: code-to-code transfer does not reveal a single dominant pivot, but a broad bridge tier, whereas English-to-code transfer concentrates more sharply in Ruby and Python (Figure~\ref{fig:functional-pivot-transfer-type}). Thus, Python's special role is better understood as English-facing affinity than as universal geometric centrality. Code-internal transfer, by contrast, relies on a broad tier of bridge languages rather than a single programming-language hub.

The main contributions of our paper are:
\begin{itemize}
    \item We show that raw cosine-based PL geometry is strongly confounded by anisotropy and code length, motivating controlled comparisons using mean-centering and length matching.
    \item We characterize controlled code--code and code--English geometry across multilingual code models, finding structured but relation-dependent PL organization rather than a universal programming-language center.
    \item We introduce pivoted retrieval as a functional diagnostic for evaluating candidate pivot languages, showing that code-to-code and English-to-code transfer favor different pivot languages.
\end{itemize}

\section{Related Work}

\paragraph{Pivot Languages in Multilingual LMs}
Recent interpretability work suggests that multilingual LMs internally route through English-like representations regardless of input language. \citet{wendler2024llamas} show that intermediate-layer representations of non-English prompts decode into English tokens with high probability; \citet{dumas2025separating} strengthen this with causal activation patching, demonstrating language-agnostic but English-aligned mid-layer concepts.
This centrality likely reflects English dominance in pretraining rather than a privileged status of English itself~\citep{zhong2025language}. \citet{he2025semantic} find that multilingual models route through shared intermediate tokens when source–target co-occurrence is low, while \citet{alabi2024hidden} show that even adapted models evolve predictions in the base model's source language until final layers. Together, these results suggest pivot structure follows from training data composition.
This structure extends beyond natural language: \citet{wu2025semantic} propose that multilingual models share a single semantic hub across modalities including code, showing that interventions on English representations affect code generation.

\paragraph{Cross-lingual Analysis of Code LMs}
A small but growing body of work analyzes how code LMs represent multiple programming languages internally. \citet{kargaran2025programming} extend pivot-language analysis to code, applying logit lens~\cite{belrose2023eliciting}, MEXA cross-lingual alignment~\cite{kargaran2025mexa}, and LAPE neuron probing~\cite{tang2024language} to two Llama-family models. They argue that English remains the implicit pivot in code LMs and that C\# aligns most closely with other PLs, with C-family languages forming a tight internal cluster that resists language-specific neuron isolation. Notably, Python ranks as the least aligned programming language with other PLs under their measure.

\citet{yin2025neuron} build on this with a probing methodology that localizes cross-PL structure in depth, identifying a band of mid-network layers where cross-PL semantic invariance peaks. \citet{yun2025beyond} approach the question from outside the model, using pre-trained embedding models on synthesized parallel code to cluster 19 PLs into paradigm-aligned families. Their C-family cluster matches the grouping reported by~\citet{kargaran2025programming}, and they additionally identify Go as the geometric centroid of their embedding space.

While the geometric analyses above point toward C-family languages and Go as central in code representation space, behavioral and training-dynamics evidence highlights a different language: Python. \citet{moumoula2026programming} document programming language confusion in code LMs: when models generate code in unintended languages despite explicit instructions, Python emerges as a ``universal attractor'', which they attribute to Python's overrepresentation in training corpora. \citet{yang2026scaling} demonstrate this asymmetry at the pre-training level: mixing Python as an auxiliary language during pre-training consistently improves performance on other target languages, while Python itself does not benefit from auxiliary mixing.

These findings are in tension: geometric analyses place Python far from other PLs and emphasize C-family centrality, while behavioral evidence points toward Python as uniquely influential. This motivates our study: we revisit cross-programming-language representations under controls for anisotropy and code length, and separately examine code–code structure and code–English alignment.

\section{Experimental Setup}

\subsection{Models.}
We study three code LMs of different architectures: \textbf{StarEncoder}~\cite{listarcoder} (145M, encoder-only), \textbf{CodeLlama-7B}~\cite{roziere2023code} (decoder-only), and \textbf{Qwen2.5-Coder-7B}~\cite{hui2024qwen2} (decoder-only, base variant). All three are pretrained on multilingual code corpora covering the programming languages we analyze. We extract mean-pooled representations at every transformer layer (comprising 12, 33, and 29 layers respectively). 

\subsection{Datasets}

We construct our evaluation dataset from the xCodeEval benchmark~\cite{khan2024xcodeeval}, a large-scale multilingual code understanding and generation dataset derived from competitive programming submissions. The dataset contains approximately 1.29 million submissions spanning 11 programming languages, with each submission linked to a unique problem identifier. We choose xCodeEval because our cross-language analyses require functionally equivalent solutions to shared problems across many programming languages, together with execution validation and paired natural-language problem descriptions.

We filter for problems that have accepted solutions in at least 7 of the 11 supported programming languages, ensuring sufficient cross-lingual coverage for alignment evaluation. For each (problem, language) pair, we retain a single canonical solution: the shortest source code with a passing execution outcome. Retaining one solution per problem–language pair avoids unequal weighting from differing numbers of accepted submissions, while selecting the shortest provides a deterministic canonicalization rule.
The resulting canonical set contains 7{,}735 solutions across 11 languages.

The 11 programming languages, ordered by number of canonical solutions, are: C++ (961), Python (959), Java (922), C\# (919), C (892), Go (701), Rust (626), Kotlin (477), Javascript (455), PHP (454), and Ruby (369). In addition, we treat the natural-language problem description as a 12th modality (``English''), giving each problem a textual representation alongside its code solutions (961 total unique problem descriptions). 


\section{Methods}

\subsection{Anisotropy Measurement}

Representational anisotropy refers to transformer embeddings occupying a narrow space rather than being uniformly distributed, causing even unrelated inputs to receive high cosine similarity. \citet{DBLP:conf/esann/DieraGS25} show this also affects code language models, making raw cosine unreliable in our setting. Following \citet{ethayarajh2019contextual}, we quantify anisotropy per layer with two complementary measures. \paragraph{Random-pair cosine} computes mean cosine similarity over $N=10{,}000$ randomly paired representations from the pooled code set:
$S_{\text{rand}_\ell} = \frac{1}{N} \sum_{i=1}^{N} \cos(h_\ell(x_i), h_\ell(y_i))$,
where $h_\ell(x)$ is the mean-pooled representation of sample $x$ at layer $\ell$. Values near one indicate severe anisotropy; zero indicates isotropy. \paragraph{Mean-norm ratio} computes the ratio of the norm of the mean representation to the mean of representation norms:
\begin{equation*}
\text{MNR}_\ell = \frac{\| \frac{1}{M}\sum_{i=1}^{M} h_\ell(x_i) \|}{\frac{1}{M}\sum_{i=1}^{M} \| h_\ell(x_i) \|}
\end{equation*}
where $M$ is the total number of code samples. Values near one indicate a dominant shared direction; near zero indicates uniform spread.

\subsection{Length Matching}
\label{sec:methods_length}

\begin{figure}[h]
    \centering
    \includegraphics[width=0.48\textwidth]{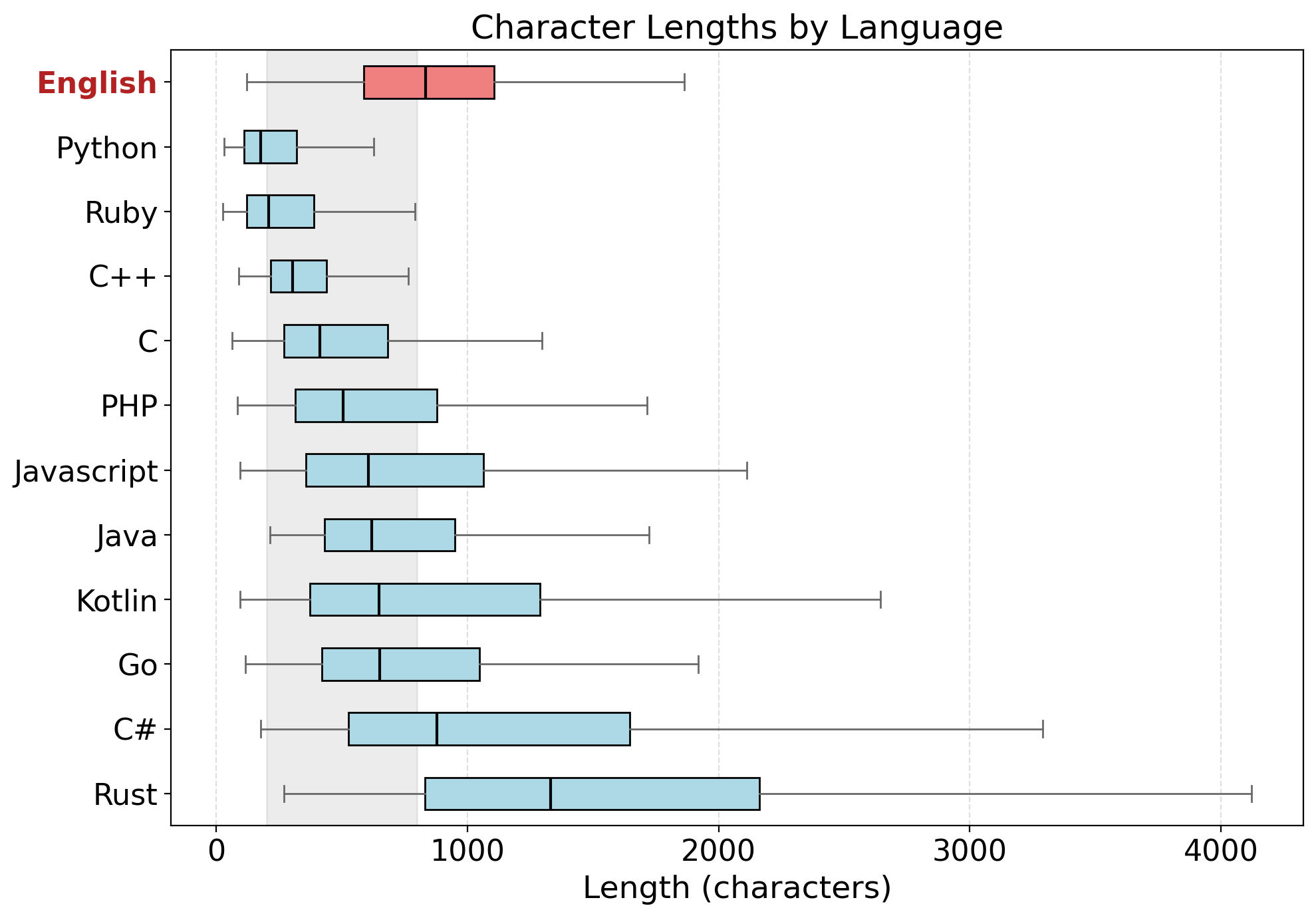}
    \caption{Box-and-whisker plot of character lengths by language.
    English refers to problem descriptions, while all other languages
    refer to the source-code solutions. The shaded region indicates
    the 200--800 character matching window used in subsequent
    analyses.}
    \label{fig:length}
\end{figure}

Programming languages vary substantially in code length even for solutions to identical problems. Figure~\ref{fig:length} shows the distribution of character counts per language in our dataset: mean lengths range from approximately 280 characters (Python) to 1700 (Rust), a six-fold difference. Since mean-pooling averages over
different numbers of token positions, longer solutions may yield representations with different variance or stronger averaging effects than shorter solutions. As a result, cosine-based similarity measures may reflect length-related representation structure rather than semantic alignment. We quantify this confound by measuring Spearman correlation between cosine-based alignment scores and the corresponding character-length statistics on the unmatched data.

To control for the length confound, we restrict cross-language analyses to
samples of length 200--800 characters. This window compresses
the six-fold mean-length spread to 1.61$\times$ within the
matched subset (371--596 chars), while retaining at least 146
samples per language. We chose this window to prioritize length
equalization over sample retention: wider windows yield modestly more samples but reintroduce length variation by including the right tails of longer-coded languages. We report the full window-selection analysis and per-language counts in Appendix~\ref{app:window}.

\subsection{Geometry Analysis}
\label{sec:methods_geometry}

To control for the two confounds described above, we use the length-matched subset and mean-center representations to address anisotropy. We define $h(x) = h_{\text{raw}}(x) - \bar{h}_{\text{center}}$, where $h_{\text{raw}}(x)$ is the mean-pooled hidden state and $\bar{h}_{\text{center}}$ is its mean over the comparison pool: all code samples for programming-language comparisons, all code and English samples for code--English comparisons.

We employ two complementary measures of cross-language alignment, chosen to estimate different alignment quantities. Centroid cosine summarizes language-level structure by comparing the mean directions of programming-language populations, while pointwise cosine estimates distribution-level affinity between code and English descriptions.
For all geometric analyses, we report results at the layer nearest normalized depth $0.5$, treating this as a representative mid-network slice. This follows prior findings that cross-programming-language semantic invariance peaks in intermediate layers~\citep{yin2025neuron}; Appendix~\ref{app:layer-selection} further verifies this choice with cross-PL contrast analysis and semantic probing.

\paragraph{Centroid cosine.} For comparing programming languages to one another, we report the cosine similarity between language centroids. Letting $\mu^a = \frac{1}{|L_a|} \sum_{x \in L_a} h(x)$ denote the centroid of language $L_a$:
\begin{equation*}S_{\text{cent}}(L_a, L_b) = \cos(\mu^a, \mu^b)\end{equation*}
This summarizes each programming language as a single point in representation space and quantifies the angular relationship between language-level directions. We use this measure for code--code comparisons because all programming-language populations are drawn from the same underlying task distribution, making language-level centroid comparisons a natural summary of cross-language structure.

\paragraph{Pointwise cosine.}
For characterizing how a programming language relates to natural-language English problem descriptions, we compute the mean pairwise cosine similarity over random code--English pairings:
\begin{equation*}
S_{\text{point}}(L_a) =
\frac{1}{N |L_a|}
\sum_{i=1}^{N}
\sum_{x^a \in L_a}
\cos\!\left(h(x^a), h(e_{i,x})\right)
\end{equation*}
where $e_{i,x}$ is an English problem description sampled uniformly from the English pool for code sample $x^a$ in random-pairing round $i$. This estimates the expected cosine similarity between code samples in language $L_a$ and English problem descriptions, providing a distribution-level measure of which programming language's representations are closest to the English-description modality in the model's embedding space. We use the pointwise formulation for code--English comparisons because code and English form cross-modal populations whose relationship may not be fully captured by a single centroid-to-centroid comparison. We use $N = 200$ random-pairing rounds per language.

\subsection{Pivoted Retrieval}
\label{sec:methods-pivoted-retrieval}

The geometric analyses above describe where programming languages lie in representation space, but do not test whether a language functions as a useful intermediate representation. We therefore introduce a \emph{pivoted retrieval} diagnostic: given a candidate pivot language $P$, how well does aligning representations through $P$ preserve same-problem retrieval? Importantly, the intermediate pivot space is explicitly imposed by our diagnostic: source and target representations are mapped into $P$ by learned alignment maps. Performance therefore measures how well $P$ functions as an intermediate coordinate system for preserving semantic correspondence, not whether the model internally routes through $P$ during inference.

We construct an evaluation set of 100 fully parallel (functionally equivalent) samples across nine languages: Python, C++, Java, C\#, C, Go, Rust, JavaScript, and PHP. Ruby and Kotlin are excluded from the evaluation panel due to lower full-parallel coverage, but are included as candidate pivot spaces. Unlike the geometric analyses, this diagnostic uses a fully parallel set rather than the length-matched subset, so that all source--target--pivot comparisons are performed over the same problems.

For each candidate pivot language $P$, we learn maps from every other representation space into $P$ using paired non-evaluation problems. Motivated by prior work using orthogonal transformations to align bilingual representation spaces for translation retrieval~\citep{smith2017offline}, we fit a centered orthogonal Procrustes map from language $L$ to pivot $P$:
\[
R_{L \rightarrow P}
=
\arg\min_{R^\top R = I}
\left\|
(A - \mu_L)R - (B - \mu_P)
\right\|_F\]
\[\qquad
\phi_{L \rightarrow P}(x)
=
(x-\mu_L)R_{L\rightarrow P}+\mu_P.
\]
Before fitting Procrustes maps, we reduce dimensionality via PCA; Appendix~\ref{app:pca-sweep} reports the chosen dimension and a sweep showing that pivot rankings are stable across a broad range. All maps are trained only on non-evaluation problems, using the same number of paired examples per map to control for differences in language-pair coverage.

We evaluate two pivoted retrieval settings. In \textbf{code-to-code retrieval}, for each ordered source--target pair $(S,T)$ with $S,T \in \mathcal{L}_{\mathrm{eval}}$ and $S \neq T$, we compare all pivots $P \in \mathcal{L}_{\mathrm{pivot}} \setminus \{S,T\}$. Source and target representations are both mapped into the pivot space:
\[
\phi_{S \rightarrow P}(h(x_p^S)),
\qquad
\phi_{T \rightarrow P}(h(x_q^T)).
\]
For each source problem $p$, the model ranks the 100 target-language candidates by cosine similarity; the correct item is the target solution to the same problem.
In \textbf{English-to-code retrieval}, English problem descriptions are used as queries and code solutions as candidates. For each target language $T \in \mathcal{L}_{\mathrm{eval}}$, we compare all pivots $P \in \mathcal{L}_{\mathrm{pivot}} \setminus \{T\}$ by mapping both English descriptions and target-language code into the pivot space:
\[
\phi_{E \rightarrow P}(h(e_p)),
\qquad
\phi_{T \rightarrow P}(h(x_q^T)).
\]

We report mean reciprocal rank (MRR) and within-task pivot ranks as the primary comparison. For each fixed retrieval task, all eligible pivots are evaluated on the same queries and candidates and ranked by MRR. We then average these ranks across tasks, random subsampling seeds, and models.

\section{Results}


\subsection{Anisotropy  Confound}
\label{sec:anisotropy}

Table~\ref{tab:anisotropy} reports random-pair cosine similarity and mean-norm ratio across all transformer layers. All three models exhibit severe anisotropy: even at their most isotropic layer, random-pair cosine remains at 0.74 (CodeLlama), 0.84 (Qwen2.5-Coder), and 0.94 (StarEncoder), with mean-norm ratios above 0.85 throughout. No layer approaches isotropy, consistent with prior findings that anisotropy is inherent to transformer self-attention~\citep{godey2024anisotropy}. We therefore apply mean-centering to all subsequent cosine analyses. The layerwise breakdown of the anisotropy measurement can be found in Appendix~\ref{app:anisotropy}.

\begin{table}[h]
\centering
\small
\begin{adjustbox}{width=\columnwidth}
\begin{tabular}{@{}lcc@{}}
\toprule
\textbf{Model} & \textbf{Random-pair cos} & \textbf{Mean-norm ratio} \\
\midrule
StarEncoder & 0.94--0.98 & 0.97--0.99 \\
CodeLlama & 0.74--0.97 & 0.85--0.99 \\
Qwen2.5-Coder & 0.84--0.95 & 0.92--0.97 \\
\bottomrule
\end{tabular}
\end{adjustbox}
\caption{Anisotropy in code language model representations across all layers. Random-pair cosine and mean-norm ratio are reported as min–max ranges over layers.}
\label{tab:anisotropy}
\end{table}

\subsection{Length Confound}
\label{sec:length}

Table~\ref{tab:confounds} reports min--max Spearman $|\rho|$ ranges between cosine-based alignment measures and code length across transformer layers, computed on the unmatched full-length data. Length is a moderate confound for code--code geometry: across models, language pairs with larger length differences receive lower centroid cosine similarities ($|\rho|=0.24$--$0.49$). The effect is much stronger for code--English alignment in decoder models, where per-language English affinity closely tracks mean code length, peaking at $|\rho|=0.88$ in CodeLlama and $|\rho|=0.96$ in Qwen2.5-Coder. StarEncoder shows little code--English length dependence (max $|\rho|=0.23$).

Length variation therefore partially distorts code--code geometry but can dominate code--English similarity in decoder architectures. We apply the 200--800 character length-matching window for all subsequent geometric analyses. Most layer-wise correlations are significant at $p<0.05$; we report the full layer-wise curves and significance markers in Appendix~\ref{app:length_corr}.

\begin{table}[h]
\centering
\small
\begin{tabular}{@{}lcc@{}}
\toprule
\textbf{Model} & \textbf{Code vs code} & \textbf{Code vs English} \\
\midrule
StarEncoder    & 0.31--0.40 & 0.01--0.23 \\
CodeLlama      & 0.33--0.47 & 0.29--0.88 \\
Qwen2.5-Coder  & 0.24--0.49 & 0.21--0.96 \\
\bottomrule
\end{tabular}
\caption{Spearman correlation $|\rho|$ min-max ranges between cosine-based alignment measures and code length, across post-embedding layers
per model. All correlations are statistically significant ($p<0.05$) except StarEncoder's code-vs-English.}
\label{tab:confounds}
\end{table}

\subsection{Cross-PL Alignment}
\label{sec:code_vs_code}

\begin{figure*}[t]
    \centering
    \includegraphics[width=\textwidth]{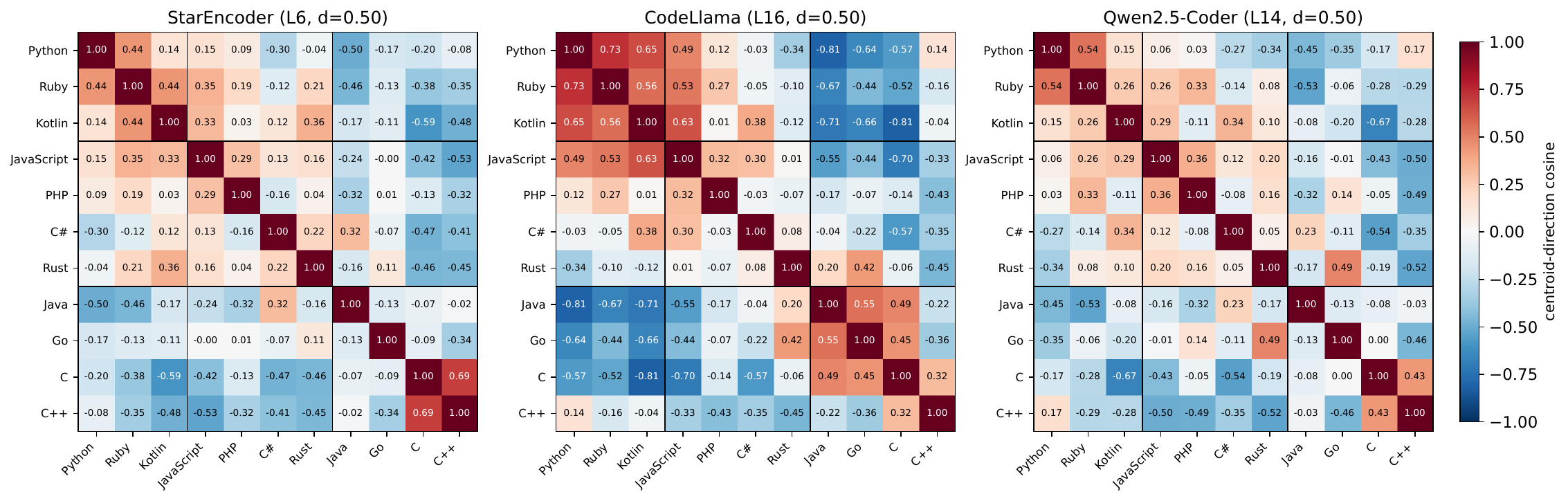}
    \caption{
    Pairwise code--code geometry at a fixed normalized depth of $0.5$ across models.
    Each heatmap reports cosine similarity between mean-centered programming-language centroid directions, computed on the length-matched code subset. The resulting geometry is structured but model-dependent, with no single programming language consistently central across models.}
    \label{fig:code-code-depth-05}
\end{figure*}

Having established that raw cosine similarities are strongly affected by anisotropy and code length, we now revisit the code--code geometry under both controls.
Figure~\ref{fig:code-code-depth-05} shows pairwise centroid-direction similarities at normalized depth $0.5$. The controlled geometry is clearly structured, but the structure is not identical across architectures. Python and Ruby consistently occupy nearby directions, suggesting a stable high-level/dynamic-language association. C and C++ also form a strong pair in StarEncoder and Qwen2.5-Coder, although this relation is weaker in CodeLlama. More broadly, CodeLlama exhibits the most polarized geometry, separating Python/Ruby/Kotlin/Javascript-like directions from Java/Go/C-like directions, whereas Qwen2.5-Coder shows a smoother organization with less extreme positive and negative values.

These results complicate the view that code representations are organized around a single universal pivot programming language. After controlling for anisotropy and length, we do observe language-family structure, but its exact form is model-dependent. In particular, C-family proximity is not uniform across architectures, and high-level languages such as Python and Ruby can be as geometrically coherent as lower-level language families.

\subsection{English Alignment}
\label{sec:code_vs_eng}

We next examine how each programming language aligns with natural-language English problem descriptions under the same length-matched, mean-centered setup. For each language, we compute pointwise random-pair cosine affinity between its code samples and randomly paired English descriptions, and rank the eleven languages within each model at normalized depth 0.5. As a surface baseline, we also rank languages by token-based Jaccard overlap between code tokens and English-description tokens.

\begin{figure}[h]
    \centering
    \includegraphics[width=0.99\linewidth]{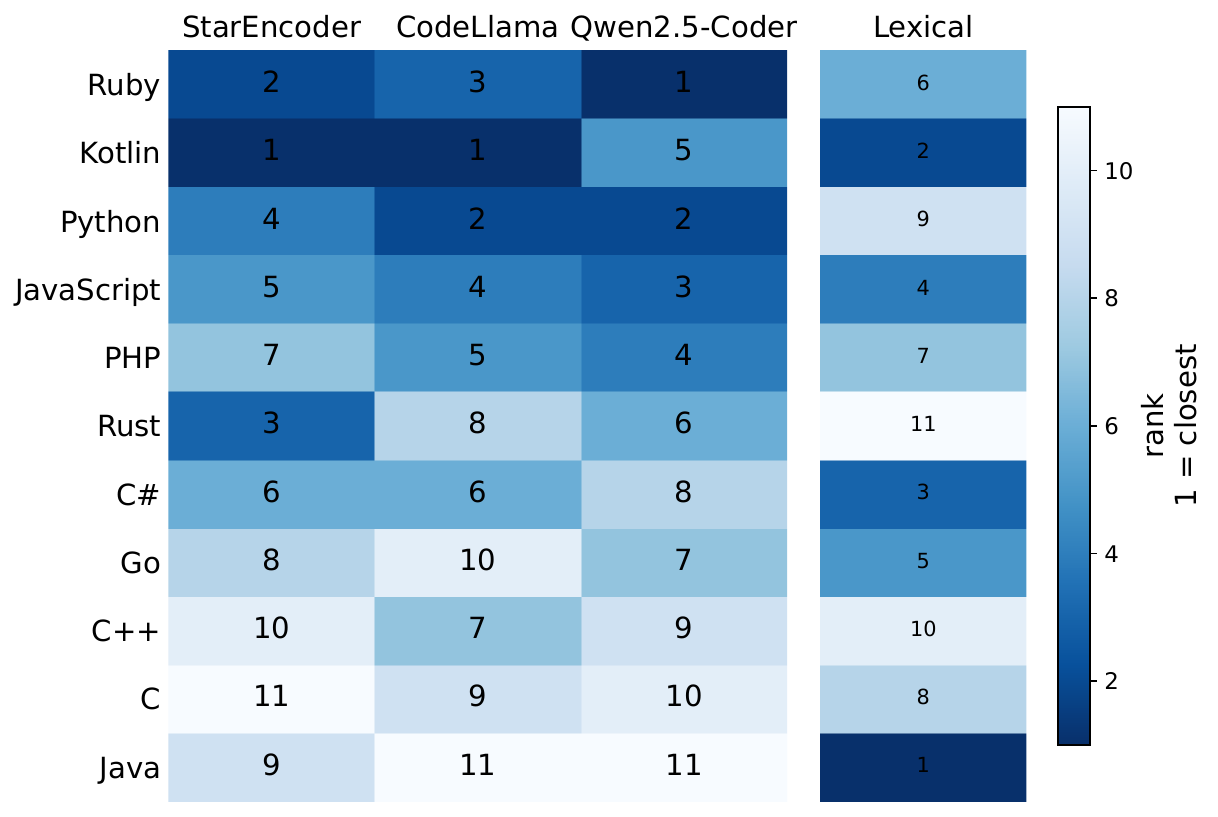}
    \caption{
Code–English alignment ranks at normalized depth 0.5. Model columns rank length-matched, mean-centered cosine similarity between each programming language and English problem descriptions. The Lexical column ranks raw Jaccard token overlap. Lower rank = closer to English. High-level scripting-like languages align most closely with English, a pattern not explained by lexical overlap alone.}
    \label{fig:code-english-alignment}
\end{figure}

Figure~\ref{fig:code-english-alignment} shows the resulting ranks. Ruby, Kotlin, Python, and JavaScript consistently occupy the top of the representation rankings, while Java, C, C++, and Go occupy the bottom. This ordering is more cross-model consistent than the code--code geometry in Section~\ref{sec:code_vs_code}: all three models agree that high-level or scripting-like languages are closest to English, while systems-oriented languages are farthest.

The lexical baseline only partially matches the alignment rankings. Java has the highest token overlap with English but ranks at the bottom in representation space, whereas Python and Ruby are strongly English-facing despite lower lexical-overlap ranks. This dissociation holds even after removing common stopwords and PL keywords (Appendix \ref{app:english_alignment_values}), confirming that token overlap does not explain the representation-level ordering.

\subsection{Pivoted Retrieval}
\label{sec:pivot}

We next ask which programming languages serve as useful intermediate spaces for retrieval. Table~\ref{tab:functional-pivots} shows that pivot behavior splits by transfer type. For code-to-code retrieval, no single pivot language dominates. PHP obtains the best average within-task pivot rank (4.36), but C++, JavaScript, Ruby, and Java follow closely, and JavaScript achieves the highest absolute MRR (0.221). Thus, code-internal transfer is better characterized as a broad bridge tier than as evidence for a unique code pivot. This tier is also heterogeneous: it includes scripting/dynamic languages such as PHP and JavaScript, but also C-family languages such as C++ and Java. English-to-code retrieval is more sharply organized. Ruby and Python obtain the best average ranks (2.43 and 3.23) and the highest MRRs (0.141 and 0.124), followed by JavaScript, C++, and PHP. Lower-ranked pivots include Go, C\#, and Rust. This mirrors Section~\ref{sec:code_vs_eng}: high-level scripting-like languages are closest to English problem descriptions.

Overall, the retrieval results show that the organizational roles attributed to a single pivot language in prior work are distributed across different
languages depending on the relation being bridged. Code-internal transfer
favors a broad, heterogeneous bridge tier, whereas English-grounded transfer concentrates more sharply in Ruby/Python-like English-facing languages. The no-pivot retrieval baseline provides complementary context. Raw code-to-code retrieval exhibits heterogeneous pairwise
structure rather than a single central language, while raw English-to-code retrieval is strongest for target languages different from the best
pivot spaces. Moreover, pivoted English-to-code retrieval achieves higher MRR than direct no-pivot English-to-code retrieval, indicating that the
learned pivot-space bottleneck can improve cross-modal same-problem
matching rather than merely reflecting which target languages are easiest to retrieve directly.
(Appendix~\ref{app:raw-retrieval-breakdown}).

\begin{table}[h!]
\centering
\small
\setlength{\tabcolsep}{4pt}
\begin{adjustbox}{max width=\linewidth}
\begin{tabular}{lrr|lrr}
\toprule
\multicolumn{3}{c|}{Code$\rightarrow$Code} &
\multicolumn{3}{c}{English$\rightarrow$Code} \\
Pivot & Rank $\downarrow$ & MRR $\uparrow$ &
Pivot & Rank $\downarrow$ & MRR $\uparrow$ \\
\midrule
PHP        & \textbf{4.36} & \textbf{0.208}$_{.077}$ & Ruby       & \textbf{2.43} & \textbf{0.141}$_{.055}$ \\
C++        & \textbf{4.39} & 0.204$_{.075}$ & Python     & \textbf{3.23} & \textbf{0.124}$_{.044}$ \\
JavaScript & \textbf{4.43} & \textbf{0.221}$_{.078}$ & JavaScript & 4.31 & 0.109$_{.030}$ \\
Ruby       & 4.55 & \textbf{0.217}$_{.091}$ & C++        & 4.63 & 0.107$_{.034}$ \\
Java       & 4.61 & 0.198$_{.071}$ & PHP        & 4.84 & 0.102$_{.026}$ \\
Go         & 4.81 & 0.199$_{.069}$ & Kotlin     & 6.08 & 0.095$_{.024}$ \\
C          & 5.01 & 0.198$_{.073}$ & C          & 6.51 & 0.092$_{.024}$ \\
Kotlin     & 5.02 & 0.203$_{.076}$ & Java       & 6.77 & 0.091$_{.025}$ \\
C\#        & 5.08 & 0.200$_{.071}$ & Go         & 6.83 & 0.090$_{.023}$ \\
Python     & 5.90 & 0.197$_{.084}$ & C\#        & 7.10 & 0.089$_{.024}$ \\
Rust       & 6.97 & 0.187$_{.060}$ & Rust       & 8.09 & 0.084$_{.021}$ \\
\bottomrule
\end{tabular}
\end{adjustbox}
\caption{
Pivoted retrieval results. Each half is sorted independently by mean within-task pivot rank; lower is better. MRR reports absolute same-problem retrieval performance, with subscripts denoting standard deviation across models, source--target tasks, and 10 subsampling seeds.
}
\label{tab:functional-pivots}
\end{table}

\paragraph{Uncertainty in pivot rankings.}

To quantify uncertainty in the pivot ordering, we use 5,000
hierarchical bootstrap replicates, resampling models and then
retrieval tasks within each sampled model
(Table~\ref{tab:pivot-tier-contrasts}). For
Code-to-code retrieval, the leading six pivots form a
broad bridge tier with an internal rank spread of 0.46
(95\% CI: [0.38, 1.43]); this tier outperforms the remaining pivots
by an average of 1.07 ranks ([0.99, 1.15]). For
English-to-code retrieval, Ruby and Python form a smaller
leading tier with an internal spread of 0.80 ranks
([0.14, 1.28]) and a substantially larger 3.29-rank separation
from the remaining pivots ([1.02, 4.54]). These results support a
broad bridge tier for code-internal transfer and a more sharply
separated English-facing tier.

\begin{table}[h!]
\centering
\small
\setlength{\tabcolsep}{3pt}
\begin{tabular}{@{}llcc@{}}
\toprule
\textbf{Setting} & \textbf{Contrast} & \textbf{Estimate} & \textbf{95\% CI} \\
\midrule
Code$\rightarrow$Code
    & Top-six spread
    & 0.46
    & [0.38, 1.43] \\
Code$\rightarrow$Code
    & Tier--remainder gap
    & 1.07
    & [0.99, 1.15] \\
English$\rightarrow$Code
    & Top-two spread
    & 0.80
    & [0.14, 1.28] \\
English$\rightarrow$Code
    & Tier--remainder gap
    & 3.29
    & [1.02, 4.54] \\
\bottomrule
\end{tabular}
\caption{
Bootstrap contrasts for pivoted retrieval.
The Code$\rightarrow$Code tier comprises PHP, C++, JavaScript,
Ruby, Java, and Go; the English$\rightarrow$Code tier comprises
Ruby and Python.
Spread denotes the maximum minus minimum mean rank within a tier.
The gap is the remainder's average rank minus the
tier's average rank.
}
\label{tab:pivot-tier-contrasts}
\end{table}

\paragraph{Robustness to instruction tuning.}
We additionally repeat the pivoted-retrieval diagnostic with Qwen2.5-Coder-7B-Instruct, using the same code and problem-description inputs. As Table~\ref{tab:instruct_mrr_gain} shows, instruction tuning improves absolute retrieval performance in both transfer settings, with a larger relative gain for English-to-code retrieval. This improvement does not remove the relation-dependent structure: code-to-code retrieval remains distributed across a broad bridge tier, whereas English-to-code retrieval keeps Ruby and Python as the top two pivots. Full instruction-tuned results are reported in Appendix~\ref{app:instruction}.

\begin{table}[h]
\centering
\small
\begin{tabular}{lccc}
\toprule
Setting & Qwen-base & Qwen-Instruct & Gain \\
\midrule
Code$\rightarrow$Code & 0.263 & 0.297 & +12.9\% \\
English$\rightarrow$Code & 0.132 & 0.164 & +24.7\% \\
\bottomrule
\end{tabular}
\caption{Instruction-tuning robustness for pivoted retrieval. Values report mean MRR over matched source--target--pivot tasks and 10 subsampling seeds.}
\label{tab:instruct_mrr_gain}
\end{table}

\section{Discussion}

\paragraph{Pivothood is relation-dependent.}
The question "What is the pivot programming language?" is under-specified. Once anisotropy and verbosity variation are controlled, we find no universal programming-language center. Instead, the representation space supports at least two distinct roles: code-internal bridge languages that preserve semantic correspondence across PLs, and English-facing interface languages that align with natural-language descriptions. This distinction reconciles previously conflicting findings. C-like surface structure organizes code--code geometry without implying that C-family languages form a universal hub; Python's behavioral prominence reflects its position on the English-facing side of code representations rather than universal centrality. Our finding that English-facing affinity and code-internal bridging favor different languages is compatible with the semantic hub hypothesis \cite{wu2025semantic}, but suggests that the hub's influence is unevenly distributed in geometry and across transfer types.

\paragraph{Bridge languages for code-internal transfer.}
Code-to-code transfer favors a broad bridge tier rather than a single dominant pivot. We interpret this not as evidence that models internally translate through any one programming language, but as evidence that some PL spaces make better intermediate coordinate systems under an explicit alignment bottleneck. The bridge tier is highly varied: PHP and JavaScript mix C-like syntax with scripting-style flexibility, C++ and Java contribute widely shared OOP structure, and Ruby brings a high-level dynamic pole. Rather than revealing a single hidden center for code, these results suggest that several programming-language spaces provide effective intermediate coordinate systems connecting different regions of the PL representations under our alignment diagnostic.

\paragraph{Python as an English-facing interface.}
Python, Ruby, Kotlin, and JavaScript are consistently closer to English descriptions than systems-oriented languages. The lexical-overlap baseline shows that this pattern is not explained by token overlap alone: Java and C\# overlap strongly with English tokens but align weakly in representation space, while Python and Ruby show the opposite pattern. We therefore interpret Python's role as English-facing rather than universally code-central. Its frequency in pretraining data, prevalence in educational contexts, and short pseudocode-like syntax may make it especially useful for English-to-code generation, explaining its behavioral prominence without requiring it to be the center of code--code geometry.

\paragraph{Practical Implications} The relation-dependent structure has several implications for multilingual code model development and evaluation. First, code-to-code and natural-language-to-code transfer should be evaluated separately rather than summarized by a single notion of cross-language centrality. Second, our results suggest that the choice of auxiliary programming languages in data mixing or curriculum design should be treated as capability-dependent: English-facing languages may be particularly relevant for natural-language grounding, whereas code-internal transfer may benefit from a broader and structurally diverse language mixture. Third, retrieval and representation-alignment systems may benefit from choosing pivot spaces according to the transfer relation, rather than relying on a single fixed pivot language across both code-to-code and English-to-code settings. More generally, claims of programming-language centrality should control for representational anisotropy and code-length variation before being interpreted as meaningful cross-language structure.

\paragraph{Future Work}
Our analyses are representational rather than causal: they characterize which languages occupy useful geometric positions but do not establish that models route through them during generation. Applying causal intervention methods such as activation patching would test whether the bridge and interface roles we identify correspond to actual computational pathways. A complementary direction is to move beyond source-code representations entirely and examine whether the same pivot structure holds over intermediate representations such as compiler IR or abstract syntax trees.

\section{Conclusion}

We revisited the question of whether multilingual code language models organize representations around a single pivot programming language, under controls for representational anisotropy and code-length variation. After applying these controls, we find that the organizational structure is relation-dependent rather than centered on any one language. Code-to-code transfer favors a broad tier of bridge languages, while English–code alignment concentrates in high-level languages such as Ruby and Python. Python's widely noted special status is better explained by its English-facing affinity than by geometric centrality. These findings suggest that the field should move from asking which language is the pivot to asking which relation a candidate language helps bridge.

\newpage

\section*{Limitations}
Our study characterizes the representational and functional organization of programming languages in code LMs: where languages sit in controlled geometry, which align with English, and which best preserve same-problem information as intermediate spaces. It does not establish that any language is causally used as a pivot during generation. Demonstrating causal pivoting would require intervention methods such as activation patching or representation steering, which we leave to future work. 

Several scope conditions further bound our findings. We study three models up to 7B parameters, and our data come from competitive-programming submissions in xCodeEval covering 11 mainstream programming languages; whether the relation-dependent structure generalizes to frontier-scale models, real-world software, or language families absent from our panel remains open. Selecting the shortest accepted solution as a deterministic canonicalization rule may also favor terse or idiomatic implementations. The pivoted-retrieval evaluation uses 100 fully parallel problems across nine languages, and Procrustes maps are fit on relatively few paired examples per language pair; we mitigate this through PCA projection and the dimensionality sweep in Appendix \ref{app:pca-sweep}, but absolute MRR values are best read alongside the within-task pivot ranks we use as the primary comparison. The length-matched subset (Appendix \ref{app:window}) over-represents the upper tail of short-coded languages and the lower tail of long-coded ones, so our geometric claims hold within a comparable length range rather than over each language's typical code. We rely on mean-pooled hidden states at fixed normalized depth 0.5; while Appendix \ref{app:layer-selection} supports this choice, alternative pooling or layer slices may shift fine-grained rankings.

\section*{Acknowledgments}

This research is co-funded by the Deutsche Forschungsgemeinschaft (DFG, German
Research Foundation) – \textit{504226141}. The authors acknowledge support by the state of Baden-Württemberg through bwHPC. L.G.P.\ acknowledges support from the MIST project, funded by the Novo Nordisk Foundation under grant reference number NNF25OC0103204.

Generative AI was used to assist with writing and editing this paper. The authors reviewed all outputs and bear full responsibility for the content.

\clearpage
\bibliography{custom}

\clearpage

\appendix

\section{Appendix}
\label{sec:appendix}
\subsection{Anisotropy}

\begin{figure}[h]
    \centering
    \includegraphics[width=\linewidth]{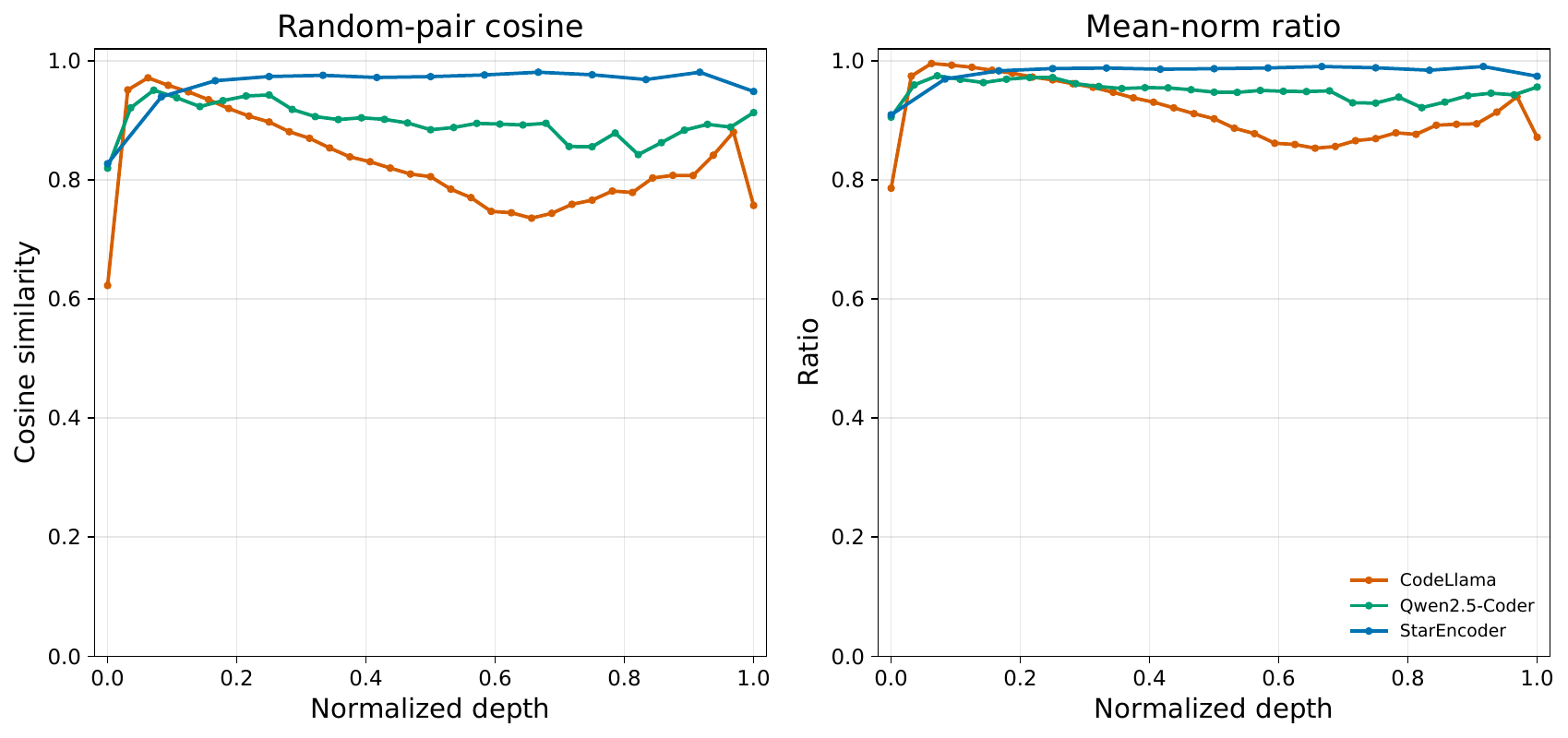}
    \caption{
    Layer-wise anisotropy across the three code language models.
    Left: mean cosine similarity between randomly paired code
    representations. Right: mean-norm ratio. Both measures remain high
    across depth, indicating strong representational anisotropy in all
    models.
    }
    \label{fig:anisotropy-layerwise}
\end{figure}

Figure~\ref{fig:anisotropy-layerwise} reports the layer-wise anisotropy
measurements summarized in Table~\ref{tab:anisotropy}. We measure
anisotropy using two complementary statistics: random-pair cosine
similarity, which estimates the average cosine similarity between
unrelated code representations, and mean-norm ratio, which measures the
strength of the shared mean direction relative to the average
representation norm.

All three models exhibit strong anisotropy across depth. StarEncoder is
the most anisotropic throughout, with random-pair cosine remaining above
$0.94$ and mean-norm ratio above $0.97$ at every layer. The decoder-only
models are less extreme but still far from isotropic: CodeLlama reaches
its lowest random-pair cosine around the middle layers, but remains above
$0.74$, while Qwen2.5-Coder remains above $0.84$. Mean-norm ratio is also
high for both decoders, staying above $0.85$ for CodeLlama and above
$0.92$ for Qwen2.5-Coder.

These results justify mean-centering before cosine-based geometric
analyses. Without this correction, cosine similarities would largely
reflect a dominant shared direction in the representation space rather
than language-specific or relation-specific alignment.

\label{app:anisotropy}

\subsection{Length Matching Window}
\label{app:window}

Our cross-language analyses use a [200, 800] character matching
window (Section~\ref{sec:length}). This appendix reports the underlying analysis and discusses limitations of the approach.

\begin{figure}[h]
    \centering
    \includegraphics[width=0.48\textwidth]{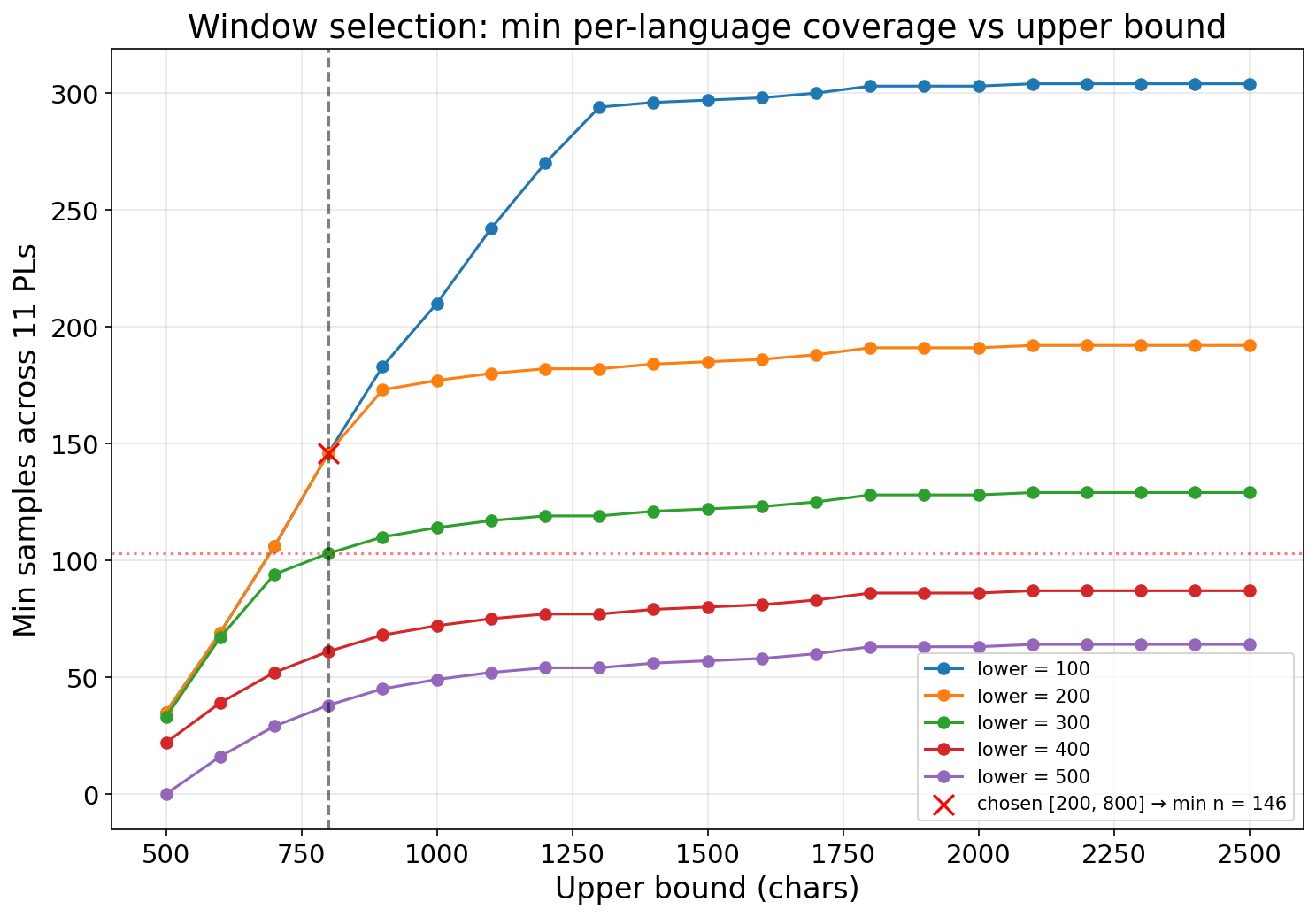}
    \caption{Minimum per-language sample count as a function of the
    upper bound, for several choices of lower bound.}
    \label{fig:window-coverage}
\end{figure}

\paragraph{Coverage analysis.}
Figure~\ref{fig:window-coverage} shows minimum per-language sample
count as a function of the upper bound, for several choices of lower
bound. Sample count grows monotonically with window width and
plateaus as each language's distributional support is exhausted. The
chosen window [200, 800] yields a minimum of 146 samples per
language, bottlenecked by Rust. Widening to [200, 1500] would gain
approximately 25\% more samples but reintroduces length variation by
including the right tails of longer-coded languages.

\paragraph{Equalization and representativeness.}
Table~\ref{tab:window-stats} reports per-language statistics within
the matched window. The matching compresses the per-language
mean-length spread from 6.1$\times$ on the full data to 1.61$\times$
within the matched subset, with tightly clustered within-language
standard deviations (128--164 chars). Residual spread is dominated
by Rust (mean = 596 chars), whose distribution is so right-skewed
that even within [200, 800] surviving samples cluster near the
upper bound. The fraction of samples retained varies substantially
across languages (0.23--0.75): the matched subset over-represents
the upper tail of shorter-coded languages (e.g., Python: 384 vs.\
279 chars on the full data) and the lower tail of longer-coded ones
(e.g., Java: 500 vs.\ 825). Findings derived from the matched
subset should therefore be interpreted as claims about
cross-language alignment within a comparable length range, rather
than about each language's typical code.

\begin{table}[h]
    \centering
    \small
    \begin{tabular}{lrrrrrr}
        \toprule
        Lang & $N_{\text{full}}$ & $N_{\text{win}}$ & frac. & $\bar{\ell}_{\text{full}}$ & $\bar{\ell}_{\text{win}}$ & $\sigma_{\text{win}}$ \\
        \midrule
        Python & 959 & 366 & 0.38 & 279 & 384 & 164 \\
        Ruby   & 369 & 166 & 0.45 & 332 & 386 & 159 \\
        C++    & 961 & 722 & 0.75 & 364 & 371 & 143 \\
        C      & 892 & 616 & 0.69 & 605 & 414 & 157 \\
        Java   & 922 & 617 & 0.67 & 825 & 500 & 149 \\
        PHP    & 454 & 293 & 0.65 & 881 & 447 & 164 \\
        Go     & 701 & 407 & 0.58 & 921 & 480 & 164 \\
        JS     & 455 & 252 & 0.55 & 995 & 467 & 163 \\
        Kotlin & 477 & 256 & 0.54 & 1055 & 455 & 159 \\
        C\#    & 919 & 416 & 0.45 & 1438 & 525 & 143 \\
        Rust   & 626 & 146 & 0.23 & 1705 & 596 & 128 \\
        \bottomrule
    \end{tabular}
    \caption{Per-language statistics within the [200, 800] matching
    window. $N_{\text{full}}$: total solutions. $N_{\text{win}}$:
    solutions in window. frac.: fraction retained. $\bar{\ell}$:
    mean length (chars). $\sigma_{\text{win}}$: standard deviation
    of length within window.}
    \label{tab:window-stats}
\end{table}

\subsection{Length Correlation}
\label{app:length_corr}

\begin{figure}[h]
    \centering
    \includegraphics[width=\columnwidth]{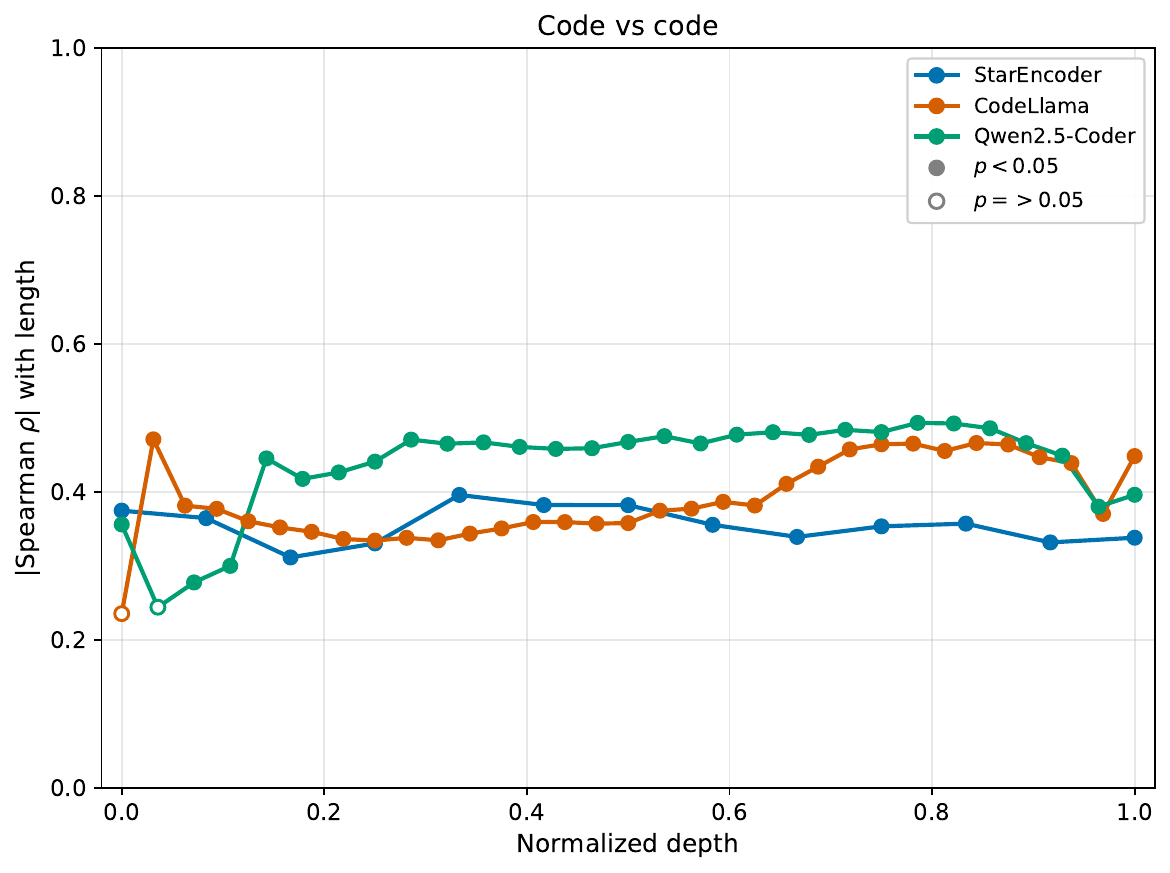}
    \caption{Layer-wise Spearman $|\rho|$ between pairwise centroid
    cosine and pairwise length difference (code-vs-code), across
    all layers per model, plotted on normalized depth
    (embedding layer at depth $0$; final layer at depth $1$).}
    \label{fig:length_corr_pw}
\end{figure}

\begin{figure}[h]
    \centering
    \includegraphics[width=\columnwidth]{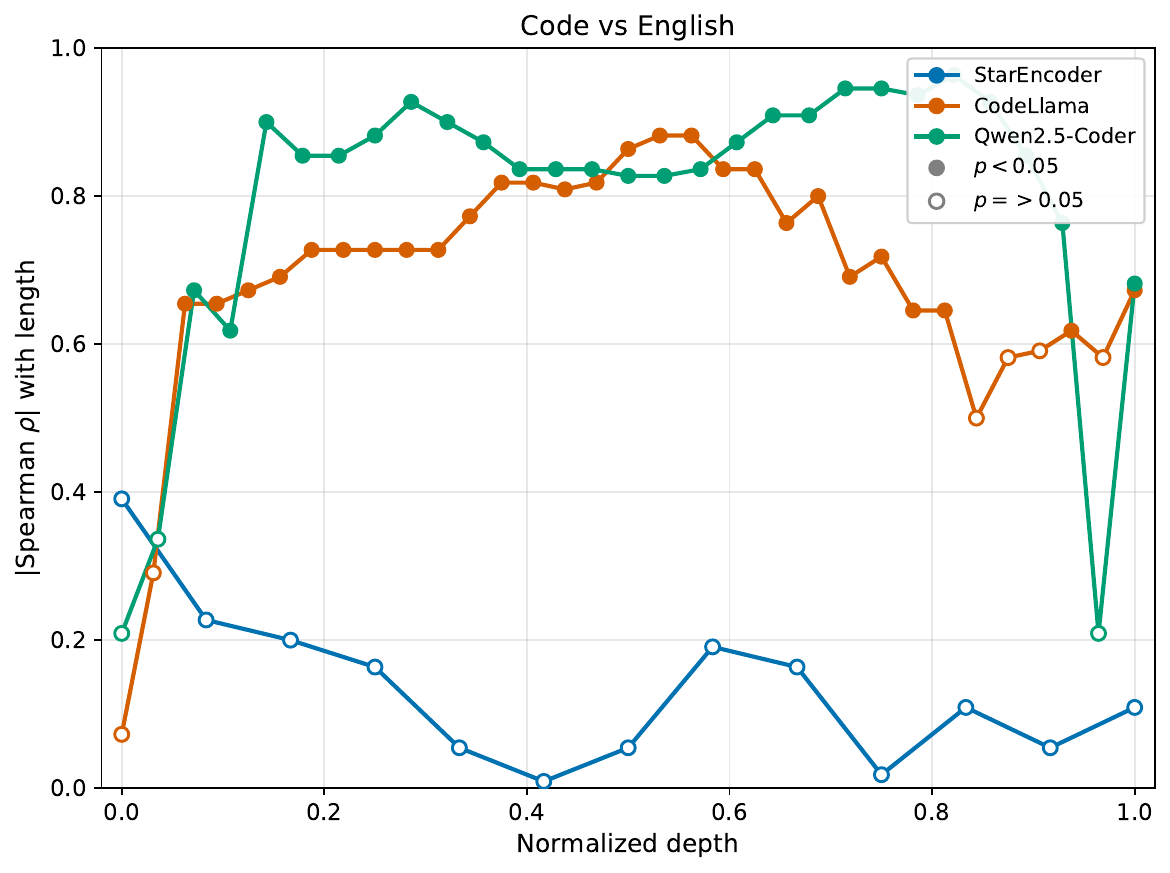}
    \caption{Layer-wise Spearman $|\rho|$ between per-language
    pointwise cosine and mean code length (code-vs-English),
    across all layers per model, plotted on normalized depth (embedding layer at depth $0$; final layer at depth $1$).}
    \label{fig:length_corr_aff}
\end{figure}

This appendix provides layer-wise detail for the Spearman correlations summarized in Table~\ref{tab:confounds}. Figures~\ref{fig:length_corr_pw} and~\ref{fig:length_corr_aff} plot $|\rho|$ across all layers (including the embedding layer L0) on the unmatched data, with filled markers indicating layers where the correlation is significant at
$p < 0.05$ and hollow markers indicating non-significant layers.

Some patterns visible in the curves are not captured by the min-max ranges. The code-vs code correlation (Figure~\ref{fig:length_corr_pw}) is stable in magnitude across depth in all three models, with only L0 in CodeLlama and L1 in
Qwen2.5-Coder failing to reach significance. The code-vs-English
correlation (Figure~\ref{fig:length_corr_aff}) shows depth structure specific to the decoders: it rises sharply from non-significant values at L0--L1 to a peak in the upper-middle layers (L17 of 33 in CodeLlama, L23 of 29 in Qwen2.5-Coder), then drifts back toward the output, with some final-layer
non-significance in both models. StarEncoder, by contrast, shows
no significant code-vs-English correlation at any layer.

\subsection{Layer Selection}
\label{app:layer-selection}

Our main geometric analyses report results at the
layer nearest normalized depth $0.5$. We choose this mid-network slice because
cross-programming-language semantic structure is expected to be most visible
after early lexical/syntactic processing but before the final layers become
more specialized to next-token prediction. To verify that this choice is not
arbitrary, we examine two complementary layer-wise diagnostics: supervised
probing and cross-programming-language semantic contrast.

\paragraph{Probing.} We train linear probes on mean-pooled hidden states from each layer. The first probe predicts multi-label semantic problem tags and is evaluated with micro-F1. This probe measures how much task-level semantic information is linearly available at each
depth. The second probe predicts the programming language identity and is
evaluated with macro-F1. This probe measures how strongly language identity is
encoded across layers.

\begin{figure}[h]
    \centering
    \includegraphics[width=\linewidth]{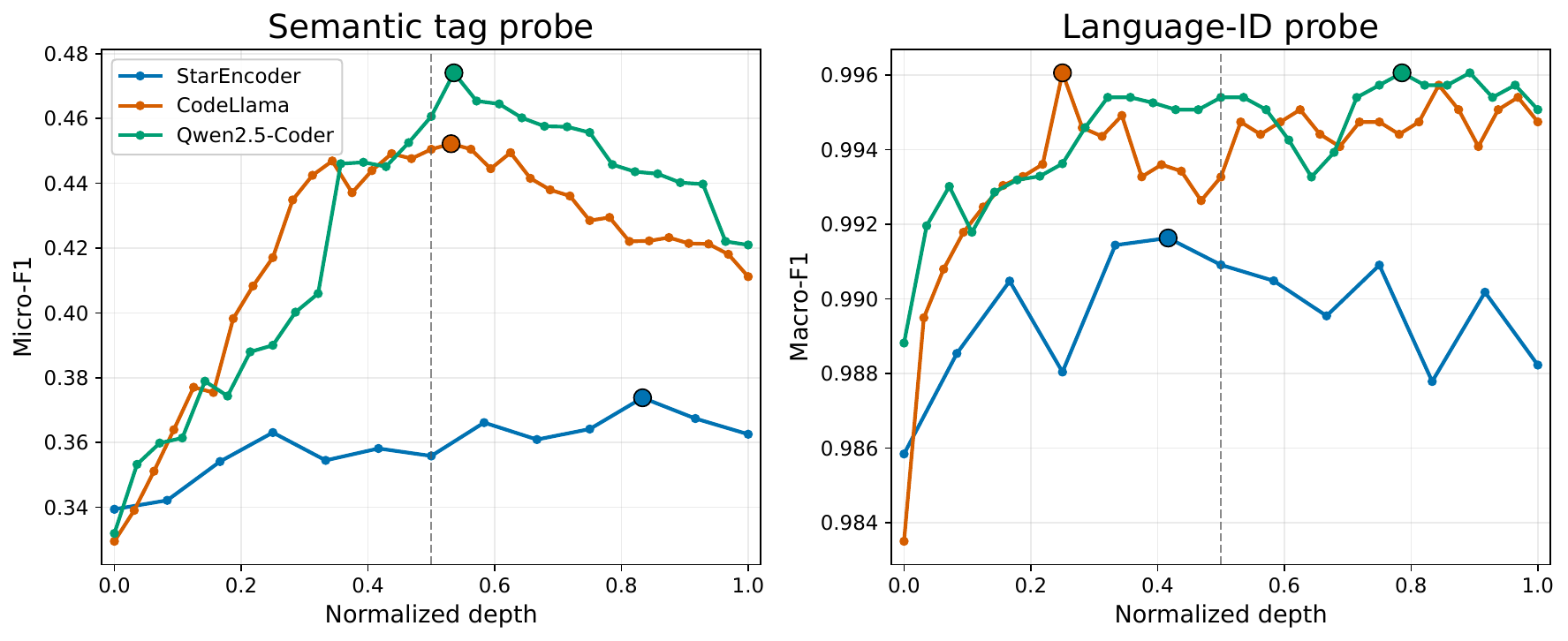}
    \caption{
    Layer-wise probing performance across models. The left panel reports
    semantic-tag prediction using micro-F1, and the right panel reports
    programming-language identification using macro-F1. The dashed vertical
    line marks normalized depth $0.5$.}
    \label{fig:probe-layer-selection}
\end{figure}

Figure~\ref{fig:probe-layer-selection} shows the resulting layer-wise curves.
For the semantic-tag probe, the decoder-only models peak close to the middle of
the network: CodeLlama and Qwen2.5-Coder both reach their highest semantic
micro-F1 around normalized depth $0.5$--$0.6$, after which performance gradually
declines. StarEncoder shows a flatter semantic curve, with weaker absolute
performance and a later peak, but its mid-layer performance remains close to
the upper range of the curve.

The language-ID probe behaves differently. Programming-language identity is
highly recoverable across almost all layers, with macro-F1 near ceiling for all
three models. Consequently, language-ID probing is less informative for
selecting a single analysis layer: it indicates that surface language identity
is available throughout the network. We therefore treat the
semantic-tag probe as the more relevant diagnostic for layer selection, and use
the language-ID probe as a sanity check that the representations still retain
programming-language information.

\subsubsection{Cross-PL Contrast}

\begin{figure}[h]
    \centering
    \includegraphics[width=0.85\linewidth]{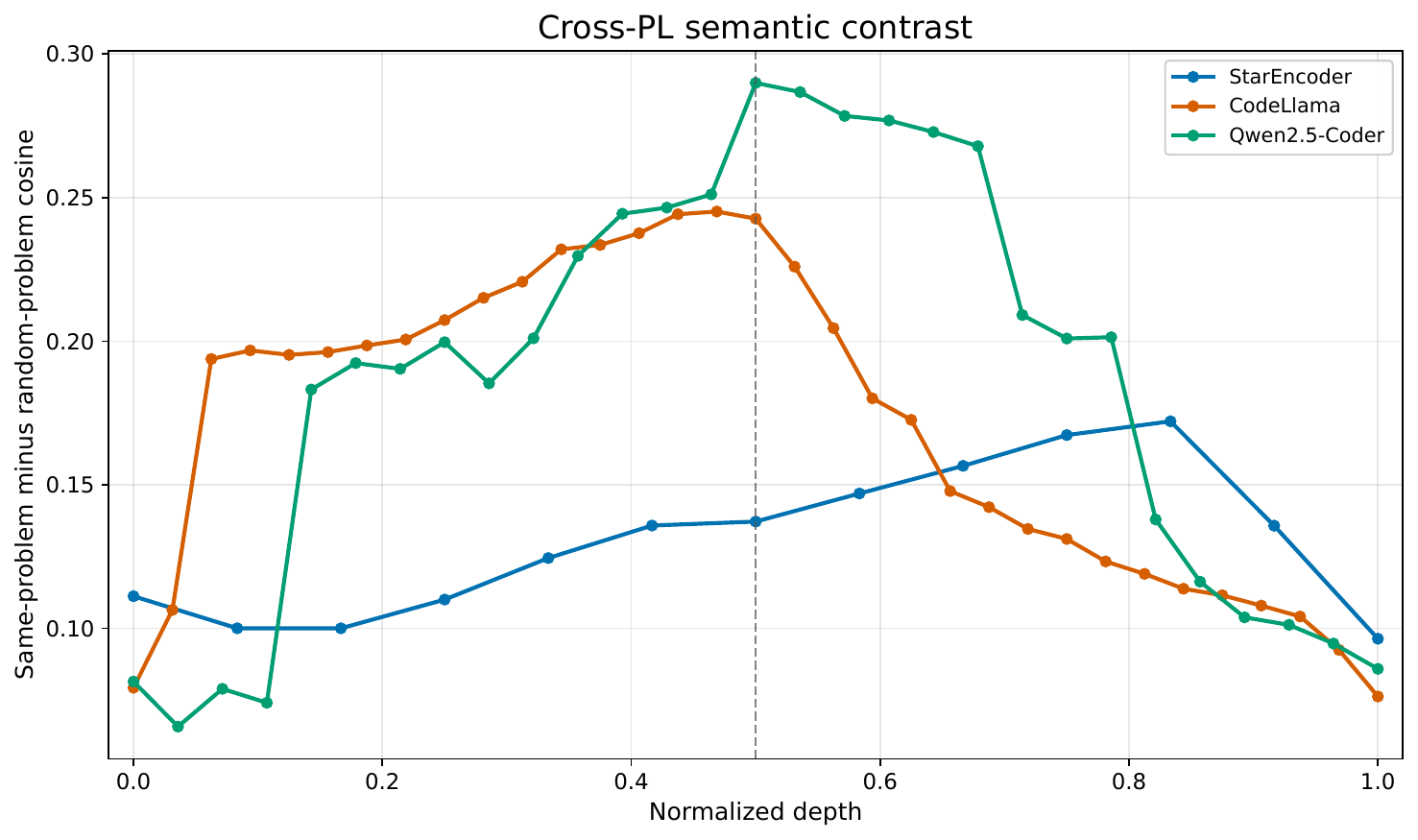}
    \caption{
    Cross-programming-language semantic contrast by layer. Higher values indicate stronger problem-level
    semantic correspondence across programming languages. The dashed vertical
    line marks normalized depth $0.5$, the representative layer used in the
    main geometric analyses.
    }
    \label{fig:cross-pl-raw-margin}
\end{figure}

As a second diagnostic, we measure where hidden states most clearly preserve
problem semantics across programming-language boundaries. For each layer, we
compare mean-centered representations of same-problem cross-language pairs
against randomly mismatched cross-language pairs over the length-matched code
set. The resulting raw contrast is the mean same-problem cosine minus the mean
random-problem cosine, averaged over usable programming-language pairs. Higher
values therefore indicate that a layer better separates shared problem semantics
from unrelated cross-language code. Figure~\ref{fig:cross-pl-raw-margin} shows
that this contrast is positive across the relevant layers and strongest around
intermediate-to-late depth, supporting the use of the layer nearest normalized
depth $0.5$ as a representative mid-network slice for the main geometry
analyses.

\subsection{Code--English Affinity Values and Lexical Baseline}
\label{app:english_alignment_values}

\begin{figure*}[h]
    \centering
    \includegraphics[width=0.95\linewidth]{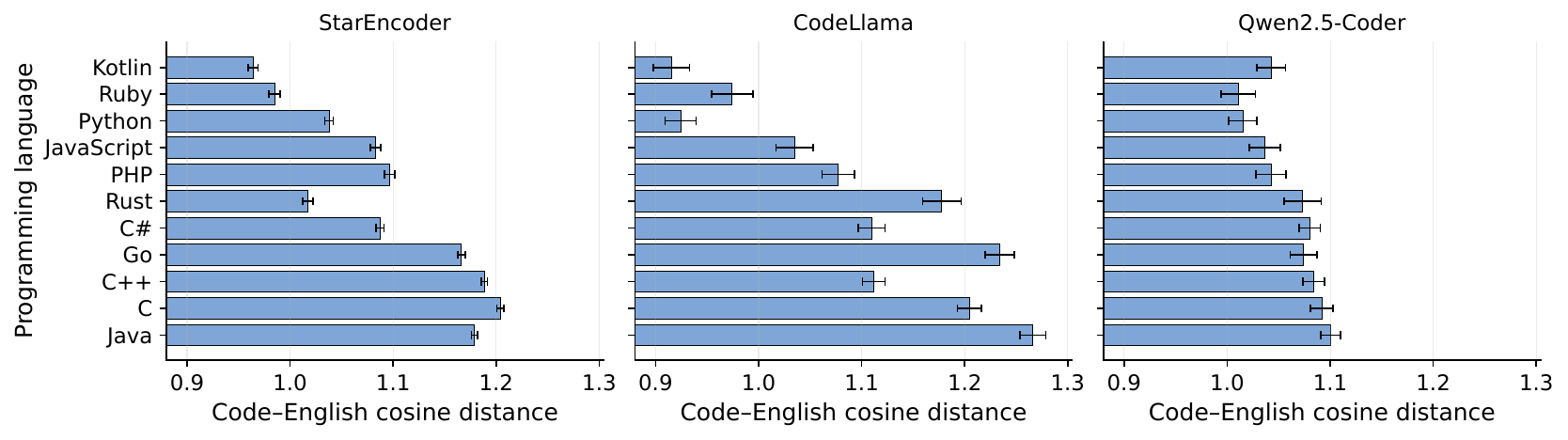}
    \caption{
    Per-model code--English cosine distance at fixed normalized depth 0.5. Bars show $1-\cos$, where cosine is the mean random-pair similarity between a programming language and English problem descriptions after length matching and mean-centering. Lower values indicate closer alignment to English. Error bars show standard deviation across 200 random-pairing rounds.
    }
    \label{fig:app-code-english-affinity}
\end{figure*}

Figure~\ref{fig:app-code-english-affinity} reports the cosine-distance values underlying the code--English ranks in Figure~\ref{fig:code-english-alignment}. We plot $1-\cos$, where cosine is the mean random-pair similarity between length-matched, mean-centered code representations and English problem-description representations. Lower values therefore indicate closer alignment to English. Across models, the raw affinity values support the same tier-level pattern as the main rank figure: high-level and scripting-like languages are closer to English, while systems-oriented languages are farther away.

\begin{figure}[h]
    \centering
    \includegraphics[width=0.99\linewidth]{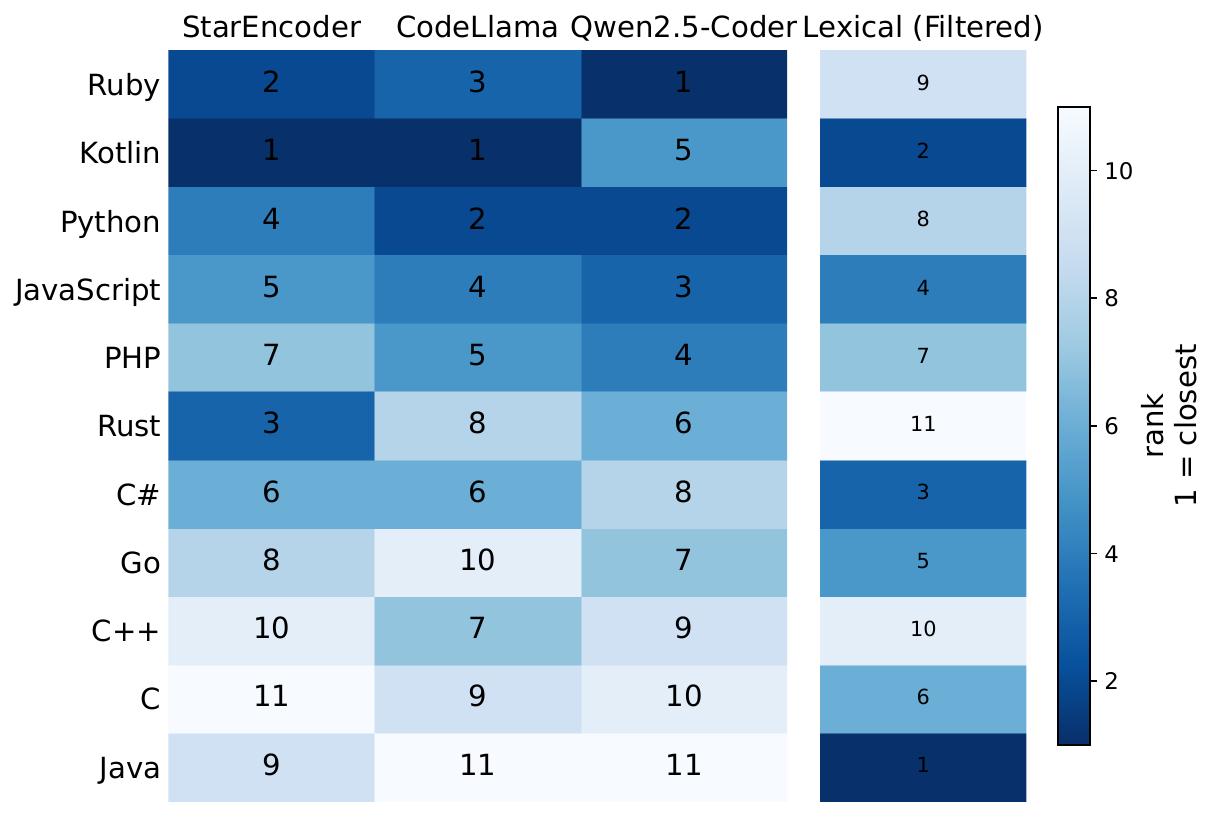}
    \caption{
    Code--English alignment ranks with a filtered token-overlap baseline. Model columns are identical to Figure~\ref{fig:code-english-alignment}. The Lexical column ranks Jaccard overlap between code tokens and English-description tokens after removing common English stopwords and programming-language keywords. Lower rank indicates closer representation-level alignment or greater lexical overlap with English.
    }
    \label{fig:app-code-english-filtered-lexical}
\end{figure}

The Lexical column in Figure~\ref{fig:code-english-alignment} is a token-based Jaccard baseline. We extract identifier-like tokens using a regular expression, split snake-case and camel-case identifiers into subtokens, lowercase all tokens, and remove numeric-only and length-one tokens. Figure~\ref{fig:app-code-english-filtered-lexical} reports a filtered variant that additionally removes common English stopwords and programming-language keywords. The filtered baseline preserves the main qualitative contrast: lexical overlap only partially matches representation-level English alignment. In particular, Java remains high under lexical overlap but low under representation alignment, while Python and Ruby remain English-facing despite comparatively low lexical-overlap ranks.

\subsection{Pivoted Retrieval PCA Sweep}
\label{app:pca-sweep}

Before fitting centered orthogonal Procrustes maps, the pivoted
retrieval diagnostic projects representations into a shared PCA
subspace. Our main experiments use $d=106$, because the balanced
code-to-pivot alignment set contains 107 paired non-evaluation
problems, so centered Procrustes has identifiable rank at most 106.
The English-to-pivot maps have more paired examples, but English-to-code
retrieval also depends on target-code-to-pivot maps; we therefore use
the same dimensionality for both retrieval settings.

To verify that the main conclusions are not artifacts of this PCA
choice, we repeat the full pivoted retrieval experiment over
\[
d \in \{16, 32, 64, 96, 106, 128, 192, 256\}.
\]
Dimensions above $106$ are included as stress tests, since they exceed
the rank supported by the balanced code alignment set. For each
dimension, we recompute pivoted retrieval across all three models and
all random subsampling seeds, then aggregate to a mean within-task pivot
rank.

Table~\ref{tab:pca-sweep-code} reports the top code-to-code pivots
across PCA dimensions. The broad bridge tier is stable, but the identity
of the single best pivot is not. Ruby is strongest at lower dimensions,
whereas PHP becomes strongest at the main dimensionality and at higher
dimensions. C++, JavaScript, and Java also remain competitive across
the sweep. This supports the interpretation that code-internal transfer
is mediated by a broad bridge tier rather than a unique pivot language.

\begin{table}[h]
\centering
\small
\begin{tabular}{r l}
\toprule
PCA dim. & Top code-to-code pivots by mean rank \\
\midrule
16  & Ruby, C++, PHP, JavaScript\\
32  & Ruby, C++, Python, JavaScript\\
64  & Ruby, C++, PHP, JavaScript\\
96  & Ruby, PHP, C++, JavaScript\\
106 & PHP, C++, JavaScript, Ruby\\
128 & PHP, JavaScript, Java, C++\\
192 & PHP, JavaScript, Go, Java\\
256 & PHP, JavaScript, Java, Go\\
\bottomrule
\end{tabular}
\caption{
Top code-to-code candidate pivots across PCA dimensions, sorted by
mean within-task pivot rank. The exact top pivot varies with PCA
dimension, but the same broad bridge tier remains competitive.
}
\label{tab:pca-sweep-code}
\end{table}

Table~\ref{tab:pca-sweep-english} reports the corresponding
English-to-code sweep. Unlike code-to-code retrieval, the top of the
English-to-code ranking is highly stable: Ruby remains the best pivot
and Python remains second across the entire sweep. JavaScript, C++,
and PHP form the next tier. This indicates that the English-facing
result is not driven by the particular choice of $d=106$.

\begin{table}[h]
\centering
\small
\begin{tabular}{r l}
\toprule
PCA dim. & Top English-to-code pivots by mean rank \\
\midrule
16  & Ruby, Python, JavaScript \\
32  & Ruby, Python, JavaScript \\
64  & Ruby, Python, JavaScript \\
96  & Ruby, Python, JavaScript \\
106 & Ruby, Python, JavaScript \\
128 & Ruby, Python, JavaScript \\
192 & Ruby, Python, JavaScript \\
256 & Ruby, Python, C++ \\
\bottomrule
\end{tabular}
\caption{
Top English-to-code candidate pivots across PCA dimensions, sorted by
mean within-task pivot rank. Ruby and Python remain the top two pivots
throughout the sweep, showing that the English-facing result is stable
under PCA dimensionality variation.
}
\label{tab:pca-sweep-english}
\end{table}

Overall, the PCA sweep strengthens the relation-dependent interpretation
of pivothood. The English-to-code setting is stable at the level of
individual top pivots, with Ruby and Python consistently leading. The
code-to-code setting is stable at the level of a bridge tier, not at the
level of a single universal pivot. We therefore interpret the main
code-to-code ordering cautiously: PHP has the best average rank at
$d=106$, but the more robust conclusion is that several languages
occupy a shared code-bridge region.

\subsection{No Pivot Retrieval}
\label{app:raw-retrieval-breakdown}

To contextualize the pivoted retrieval diagnostic, we also evaluate raw source--target retrieval without Procrustes alignment and without an intermediate pivot space. This baseline uses the same 100 fully parallel evaluation problems and the same PCA representation space as the main pivoted retrieval experiment. For code-to-code retrieval, each source-language solution is used directly as a query against the 100 target-language candidate solutions. For English-to-code retrieval, each English problem description is used directly as a query against the 100 target-language code candidates. Retrieval is evaluated by same-problem mean reciprocal rank (MRR).

\begin{figure}[h]
\centering
\includegraphics[width=\linewidth]{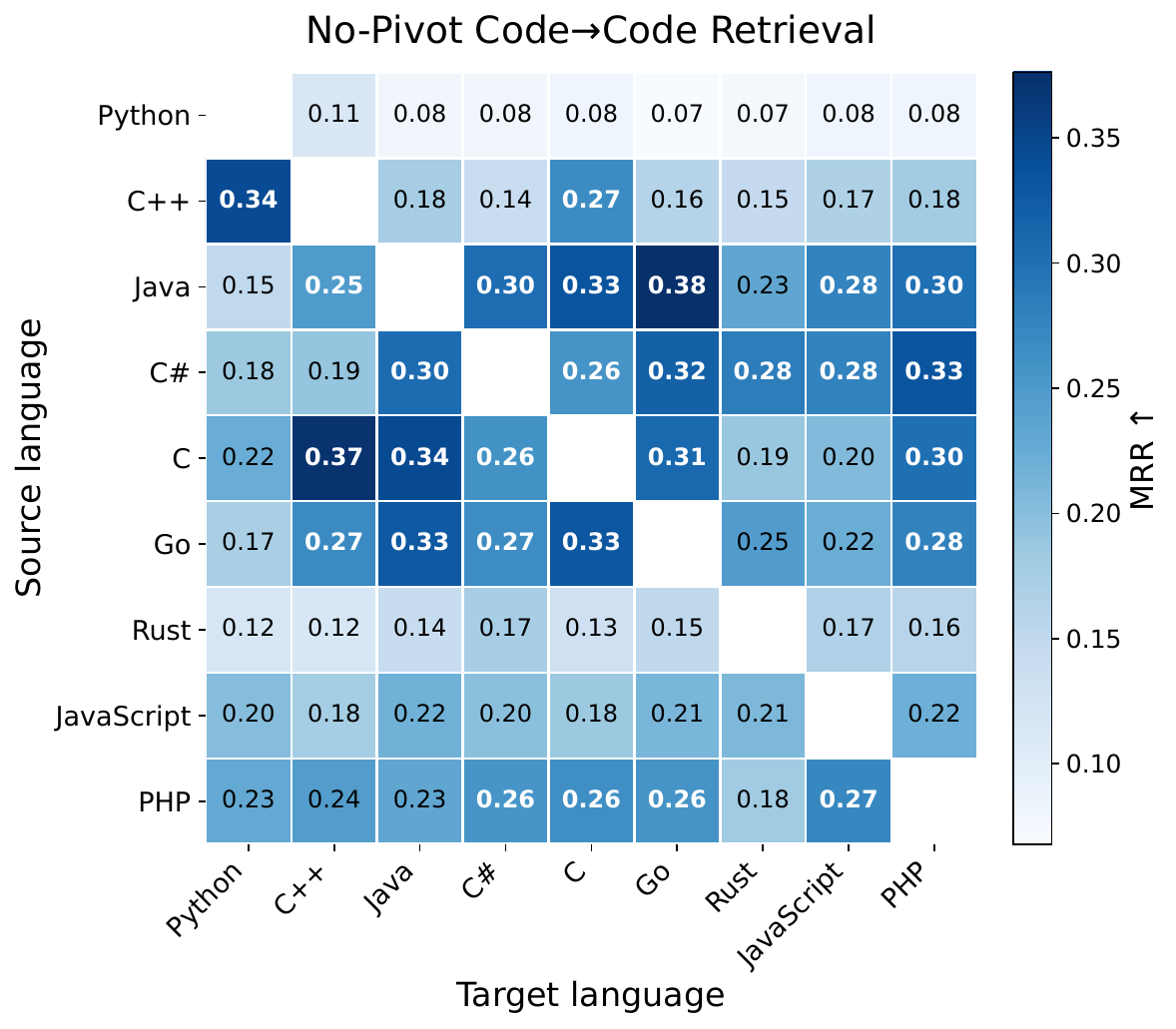}
\caption{
No-pivot code$\rightarrow$code retrieval by source and target language. Entries report mean MRR across models using the same evaluation problems as the pivoted retrieval experiment. Darker blue indicates higher retrieval performance.
}
\label{fig:raw-code-code-matrix}
\end{figure}

Figure~\ref{fig:raw-code-code-matrix} shows the resulting code-to-code MRR matrix by source and target language. The pattern is structured but heterogeneous: retrieval quality varies substantially across source--target directions, and no single programming language emerges as uniformly central across all pairs. This supports the interpretation of the pivoted retrieval results in Section~\ref{sec:pivot}: code-internal transfer is better described by multiple bridge-like relations than by a single universal code pivot.

\begin{figure}[h]
\centering
\includegraphics[width=\linewidth]{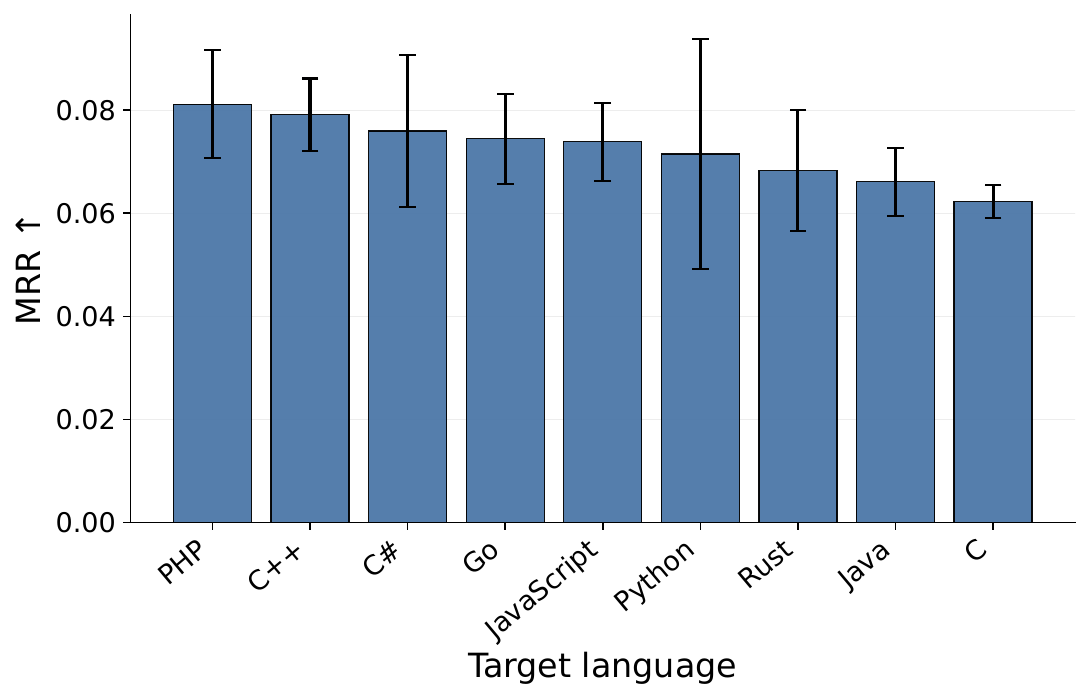}
\caption{
No-pivot English$\rightarrow$code retrieval by target language. Bars report mean MRR across models using the same evaluation problems as the pivoted retrieval experiment; error bars denote standard deviation across models.
}
\label{fig:raw-english-code}
\end{figure}

Figure~\ref{fig:raw-english-code} shows the corresponding raw English-to-code retrieval results by target language. Unlike the pivoted retrieval diagnostic, raw English-to-code retrieval does not concentrate around Ruby or Python; instead, the strongest direct target languages include PHP, C++, and C\#. This indicates that the pivoted English-to-code result is not simply a reflection of which target languages are easiest to retrieve directly from English. Rather, pivoted retrieval measures a different property: how well a candidate intermediate representation space preserves same-problem correspondence when both English queries and code candidates are mapped through that space.

\subsection{Robustness to Instruction tuning}
\label{app:instruction}

Table~\ref{tab:functional-pivots-qwen-instruct} reports the full pivoted-retrieval results for Qwen2.5-Coder-7B-Instruct. The instruction-tuned model improves absolute same-problem retrieval performance relative to the base Qwen model, especially for English$\rightarrow$code transfer in relative terms. At the same time, the qualitative relation-dependent structure remains unchanged. Code$\rightarrow$code retrieval continues to lack a single dominant pivot: the top-ranked pivots form a broad bridge tier and their ordering changes under instruction tuning. By contrast, English$\rightarrow$code retrieval remains more concentrated, with Ruby and Python staying as the top two pivots and C++/JavaScript remaining in the upper tier. These results suggest that instruction tuning strengthens the functional retrieval signal without collapsing the distinction between code-internal bridge spaces and English-facing intermediate spaces.

\begin{table}[h]
\centering
\small
\setlength{\tabcolsep}{4pt}
\begin{adjustbox}{max width=\linewidth}
\begin{tabular}{lrr|lrr}
\toprule
\multicolumn{3}{c|}{Code$\rightarrow$Code} &
\multicolumn{3}{c}{English$\rightarrow$Code} \\
Pivot & Rank $\downarrow$ & MRR $\uparrow$ &
Pivot & Rank $\downarrow$ & MRR $\uparrow$ \\
\midrule
Java       & \textbf{4.12} & 0.301$_{.075}$ & Ruby       & \textbf{2.32} & \textbf{0.203}$_{.041}$ \\
Ruby       & \textbf{4.25} & \textbf{0.313}$_{.088}$ & Python     & \textbf{3.88} & \textbf{0.184}$_{.036}$ \\
PHP        & \textbf{4.27} & \textbf{0.307}$_{.081}$ & C++        & \textbf{3.90} & \textbf{0.184}$_{.036}$ \\
Go         & 4.42 & 0.301$_{.080}$ & JavaScript & 3.98 & 0.181$_{.024}$ \\
C          & 4.54 & 0.296$_{.074}$ & C          & 4.85 & 0.168$_{.025}$ \\
Kotlin     & 4.64 & 0.305$_{.077}$ & PHP        & 5.28 & 0.161$_{.020}$ \\
C++        & 4.72 & 0.288$_{.073}$ & Kotlin     & 5.89 & 0.159$_{.022}$ \\
JavaScript & 4.78 & \textbf{0.322}$_{.074}$ & Java       & 6.11 & 0.155$_{.021}$ \\
C\#        & 5.32 & 0.293$_{.076}$ & C\#        & 7.74 & 0.142$_{.020}$ \\
Python     & 6.65 & 0.271$_{.081}$ & Go         & 8.21 & 0.136$_{.015}$ \\
Rust       & 7.61 & 0.271$_{.060}$ & Rust       & 8.70 & 0.133$_{.018}$ \\
\bottomrule
\end{tabular}
\end{adjustbox}
\caption{
Instruction-tuned pivoted retrieval results for Qwen2.5-Coder-7B-Instruct. Each half is sorted independently by mean within-task pivot rank; lower is better. MRR reports absolute same-problem retrieval performance, with subscripts denoting standard deviation across retrieval tasks and 10 subsampling seeds.
}
\label{tab:functional-pivots-qwen-instruct}
\end{table}

\end{document}